\documentclass[10pt]{article}

\usepackage[utf8]{inputenc}
\usepackage[T1]{fontenc}
\usepackage{lmodern}
\usepackage{amsmath,amssymb}
\usepackage{booktabs}
\usepackage{graphicx}
\usepackage{microtype}
\usepackage{xcolor}
\usepackage{array}
\usepackage{enumitem}
\usepackage{seqsplit}
\usepackage[margin=1in]{geometry}
\usepackage[authoryear,round]{natbib}
\usepackage[colorlinks=true,linkcolor=black,citecolor=blue!50!black,urlcolor=blue!50!black]{hyperref}
\usepackage[nameinlink,noabbrev]{cleveref}
\usepackage{url}
\newcommand{\breakhash}[1]{\texttt{\seqsplit{#1}}}

\graphicspath{{figures/}}
\setlist{nosep,leftmargin=*}
\newcommand{\VHistories}{42}% fast_result_summary.json
\newcommand{\VCells}{900}% fast_result_summary.json
\newcommand{\VLearningRate}{0.03}% fast_result_summary.json
\newcommand{\VCellsPrelock}{1{,}728}% frozen_config.json
\newcommand{\VIccNear}{0.996}% analysis_values.json:reliability
\newcommand{\VIccFar}{0.964}% analysis_values.json:reliability
\newcommand{\VIccConflict}{0.781}% analysis_values.json:reliability
\newcommand{\VIccConflictInt}{[0.656, 0.988]}% analysis_values.json:reliability
\newcommand{\VIccMin}{0.781}% analysis_values.json:reliability
\newcommand{\VIccMinInt}{[0.673, 0.983]}% analysis_values.json:reliability
\newcommand{\VConfBetween}{0.798}% f4_conflict_icc_decomposition.json
\newcommand{\VConfWithin}{0.224}% f4_conflict_icc_decomposition.json
\newcommand{\VNearWithin}{0.003}% f4_conflict_icc_decomposition.json
\newcommand{\VFarWithin}{0.021}% f4_conflict_icc_decomposition.json
\newcommand{\VIccTwoProj}{0.877}% f4_conflict_icc_decomposition.json
\newcommand{\VIccTwoProjLo}{0.792}% instrument_v2_01/result.json
\newcommand{\VIccThree}{0.940}% conflict_repeat3/result.json
\newcommand{\VIccThreeInt}{[0.903, 0.997]}% conflict_repeat3/result.json
\newcommand{\VIccThreeUnits}{24}% conflict_repeat3/result.json
\newcommand{\VRmseBzero}{3.364}% analysis_values.json:prediction
\newcommand{\VRmseBone}{1.328}% analysis_values.json:prediction
\newcommand{\VRmseBgamma}{1.387}% analysis_values.json:prediction
\newcommand{\VRmseFlex}{2.149}% analysis_values.json:prediction
\newcommand{\VRmseHfull}{1.644}% analysis_values.json:prediction
\newcommand{\VRmseState}{1.612}% analysis_values.json:prediction
\newcommand{\VRmseStwo}{1.912}% analysis_values.json:prediction
\newcommand{\VRmseSfour}{1.951}% analysis_values.json:prediction
\newcommand{\VGainState}{$-$21.4}% analysis_values.json:prediction
\newcommand{\VGainStateInt}{[$-$91.9, 8.1]}% analysis_values.json:prediction
\newcommand{\VGainHfull}{$-$23.8}% analysis_values.json:prediction
\newcommand{\VGainHfullInt}{[$-$96.6, 30.0]}% analysis_values.json:prediction
\newcommand{\VGainBgamma}{$-$16.9}% analysis_values.json:prediction
\newcommand{\VGainBgammaInt}{[$-$88.5, 13.4]}% analysis_values.json:prediction
\newcommand{\VFlexDiff}{0.537}% analysis_values.json:prediction
\newcommand{\VFlexDiffInt}{[$-$0.412, 0.989]}% analysis_values.json:prediction
\newcommand{\VKstar}{1}% analysis_values.json:prediction
\newcommand{\VLofoMean}{$-$80.3}% analysis_values.json:prediction
\newcommand{\VLofoRep}{45.7}% analysis_values.json:prediction
\newcommand{\VLofoDiv}{$-$245.7}% analysis_values.json:prediction
\newcommand{\VLofoCon}{$-$40.9}% analysis_values.json:prediction
\newcommand{\VFamLabelGain}{$-$0.03}% analysis_values.json:prediction
\newcommand{\VFitTrainBone}{0.326}% f5_hfull_overfitting.json
\newcommand{\VFitTrainHfull}{0.289}% f5_hfull_overfitting.json
\newcommand{\VFitValBone}{0.642}% f5_hfull_overfitting.json
\newcommand{\VFitValHfull}{1.185}% f5_hfull_overfitting.json
\newcommand{\VShrunkVal}{0.629}% hfull_summary_ridge (fit_lock.json, evaluation/result.json)
\newcommand{\VShrunkTest}{1.340}% hfull_summary_ridge (fit_lock.json, evaluation/result.json)
\newcommand{\VGainShrunk}{$-$0.9}% hfull_summary_ridge (fit_lock.json, evaluation/result.json)
\newcommand{\VGainShrunkInt}{[$-$2.4, 1.3]}% hfull_summary_ridge (fit_lock.json, evaluation/result.json)
\newcommand{\VCtrlOrigHalf}{0.205}% analysis_values.json:c06
\newcommand{\VCtrlOrigHalfInt}{[0.047, 0.324]}% analysis_values.json:c06
\newcommand{\VCtrlOrigHalfPos}{5/6}% analysis_values.json:c06
\newcommand{\VCtrlOrigTwo}{0.044}% analysis_values.json:c06
\newcommand{\VCtrlHalfPeak}{0.211}% f3_c06_measurement_window.json
\newcommand{\VCtrlHalfPeakStep}{85}% f3_c06_measurement_window.json
\newcommand{\VCtrlTwoEarly}{0.753}% f3_c06_measurement_window.json
\newcommand{\VCtrlTwoLate}{0.044}% f3_c06_measurement_window.json
\newcommand{\VNoiseChange}{$-$0.204}% f3_c06_measurement_window.json
\newcommand{\VNoiseChangeInt}{[$-$0.324, $-$0.137]}% f3_c06_measurement_window.json
\newcommand{\VCtrlHalf}{0.208}% instrument_v2_01/result.json
\newcommand{\VCtrlHalfInt}{[0.023, 0.304]}% instrument_v2_01/result.json
\newcommand{\VCtrlHalfPos}{9/12}% instrument_v2_01/result.json
\newcommand{\VCtrlQuarter}{0.684}% instrument_v2_01/result.json
\newcommand{\VCtrlQuarterInt}{[0.492, 0.930]}% instrument_v2_01/result.json
\newcommand{\VCtrlQuarterPos}{11/12}% instrument_v2_01/result.json
\newcommand{\VCtrlEighth}{1.447}% instrument_v2_01/result.json
\newcommand{\VCtrlEighthInt}{[1.192, 1.490]}% instrument_v2_01/result.json
\newcommand{\VCtrlEighthPos}{11/12}% instrument_v2_01/result.json
\newcommand{\VCtrlNormErr}{2.1\times 10^{-16}}% instrument_v2_01/result.json
\newcommand{\VPrefixEnd}{85}% instrument_v2_01/result.json
\newcommand{\VCollPairs}{1{,}584}% f6_collision_coverage.json
\newcommand{\VCollBaseline}{670}% f6_collision_coverage.json
\newcommand{\VCollPredicted}{1}% f6_collision_coverage.json
\newcommand{\VPowerFiveT}{1{,}035}% f7_history_gain_power.json
\newcommand{\VPowerFiveE}{723}% f7_history_gain_power.json
\newcommand{\VPowerTenT}{276}% f7_history_gain_power.json
\newcommand{\VPowerTenE}{195}% f7_history_gain_power.json
\newcommand{\VPcrejRatioNear}{38}% postselection_controls.json
\newcommand{\VPcrejRatioFar}{15}% postselection_controls.json
\newcommand{\VPcrejRatioConflict}{14}% postselection_controls.json
\newcommand{\VPcrejZThirteen}{$-$0.06}% pcrej_production_01/result.json
\newcommand{\VPcrejZFourteen}{$-$0.05}% pcrej_production_01/result.json
\newcommand{\VPcrejZFifteen}{0.39}% pcrej_production_01/result.json
\newcommand{\VPcrejZFifteenInt}{[0.16, 0.73]}% pcrej_production_01/result.json
\newcommand{\VPcrejZSixteen}{10.4}% pcrej_production_01/result.json
\newcommand{\VPcrejZSixteenInt}{[6.0, 15.3]}% pcrej_production_01/result.json
\newcommand{\VCommitRevision}{13c822f4f887}% instrument_v2_01/result.json
\newcommand{\VConfigHash}{8fc3213a7eaa6b09}% sha256 of frozen_config.json
\newcommand{\VValuesHash}{1563631c6fca8958}% sha256 of analysis_values.json
\newcommand{\MTrajectories}{1{,}560}% v2_summary.json
\newcommand{\MAtCap}{832}% v2_summary.json
\newcommand{\MFamilyWins}{8/13}% v2_summary.json
\newcommand{\MStateBoneTwelve}{13.1}% v2_summary.json
\newcommand{\MStateBoneTwelveInt}{[2.4, 23.7]}% v2_summary.json
\newcommand{\MStateCurTwelve}{30.3}% v2_summary.json
\newcommand{\MStateCurTwelveInt}{[15.8, 39.4]}% v2_summary.json
\newcommand{\MStateCurTwentyFour}{20.9}% v2_summary.json
\newcommand{\MStateCurTwentyFourInt}{[5.3, 30.9]}% v2_summary.json
\newcommand{\MStateCurFortyEight}{0.8}% v2_summary.json
\newcommand{\MStateCurFortyEightInt}{[$-$19.8, 15.2]}% v2_summary.json
\newcommand{\MFullBoneFortyEight}{15.1}% v2_summary.json
\newcommand{\MFullBoneFortyEightInt}{[0.8, 28.4]}% v2_summary.json
\newcommand{\MHundredBone}{70.0}% v2_summary.json
\newcommand{\MHundredState}{0.418}% v2_summary.json
\newcommand{\MOverflow}{12}% v2_summary.json
\newcommand{\MInterpBoneTwelve}{0.464}% v2_summary.json
\newcommand{\MInterpStateTwelve}{0.499}% v2_summary.json
\newcommand{\MInterpBoneFortyEight}{0.279}% v2_summary.json
\newcommand{\MInterpStateFortyEight}{0.288}% v2_summary.json
\newcommand{\SRuns}{72}% cifar_spectral_01_analysis_v2/analysis.json
\newcommand{\SBestSix}{0.0684}% cifar_spectral_01_analysis_v2/analysis.json
\newcommand{\SQuadSix}{0.0669}% cifar_spectral_01_analysis_v2/analysis.json
\newcommand{\SCurSix}{0.287}% cifar_spectral_01_analysis_v2/analysis.json
\newcommand{\SBestTwelve}{0.0661}% cifar_spectral_01_analysis_v2/analysis.json
\newcommand{\SQuadTwelve}{0.0492}% cifar_spectral_01_analysis_v2/analysis.json
\newcommand{\SBestTwentyFour}{0.0496}% cifar_spectral_01_analysis_v2/analysis.json
\newcommand{\SQuadTwentyFour}{0.0374}% cifar_spectral_01_analysis_v2/analysis.json
\newcommand{\SCurTwentyFour}{0.136}% cifar_spectral_01_analysis_v2/analysis.json
\newcommand{\IRuns}{56}% cifar_incremental_forecast_01/analysis.json
\newcommand{\ICurrentFirst}{41.2}% cifar_incremental_forecast_01/analysis.json
\newcommand{\ICurrentLast}{22.9}% cifar_incremental_forecast_01/analysis.json
\newcommand{\IEsnWorse}{120/120}% cifar_incremental_forecast_01/analysis.json
\newcommand{\IGruMaxGain}{0.06}% cifar_incremental_forecast_01/analysis.json
\newcommand{\IObservations}{5{,}544}% cifar_incremental_forecast_01/analysis.json

\title{When does a network's training history predict its future learning better than its current state?\\[0.3em]
\large Evidence from a response probe and a forecasting screen}

\author{%
  Martin Hofmann$^{1}$ \qquad Patrick M\"ader$^{1,2,3}$\\[0.6em]
  \small $^{1}$ Data-intensive Systems and Visualization Group (dAI.SY),
  Technische Universit\"at Ilmenau,\\
  \small Max-Planck-Ring 14, 98693 Ilmenau, Thuringia, Germany\\
  \small $^{2}$ German Centre for Integrative Biodiversity Research (iDiv)
  Halle--Jena--Leipzig,\\
  \small Deutscher Platz 5e, 04103 Leipzig, Saxony, Germany\\
  \small $^{3}$ Faculty of Biological Sciences, Friedrich Schiller University,\\
  \small F\"urstengraben 1, 07745 Jena, Thuringia, Germany\\[0.6em]
  \small \href{mailto:martin.hofmann@tu-ilmenau.de}{\texttt{martin.hofmann@tu-ilmenau.de}}
  \qquad
  \href{mailto:patrick.maeder@tu-ilmenau.de}{\texttt{patrick.maeder@tu-ilmenau.de}}\\
  \small ORCID:
  \href{https://orcid.org/0000-0002-4440-3317}{0000-0002-4440-3317}
  \qquad
  \href{https://orcid.org/0000-0001-6871-2707}{0000-0001-6871-2707}
}

\date{}

\begin{document}
\maketitle

\begin{abstract}
Networks that behave alike now can still learn differently when training continues. Work on loss of
plasticity and critical periods shows that the path to a state shapes what follows; it does not show
whether the path carries information that a measurement of the state itself misses. We ask when the
training history of a network predicts its future learning better than its current state. In a main
study, small multilayer perceptrons were trained under three history regimes (\VHistories{} histories),
and future learning was measured at four checkpoints by a short probe: a copy of the network trained for
100 updates on a new task. Before the prediction result was read, the protocol checked the probe. It
responded monotonically to a function-preserving rescaling of hidden units, repeated measurements agreed
(intraclass correlation \VIccThree, \VIccThreeInt, in the least reliable class, mean of three repeats),
and a re-initialisation of units was visible directly after it but not 100 to 200 updates later. A
history state of at most four dimensions did not improve on a calibrated model of the current state
(gain \VGainState\%, 90\% interval \VGainStateInt; required in advance: 10\%). A companion screen on
\MTrajectories{} synthetic regression runs asked the same question for a target further away, the final
error of the run. There, history models forecast better than the current validation error after 12 of up
to 240 epochs (compact state \MStateCurTwelve\%, \MStateCurTwelveInt, a contextual comparison) and were
not distinguishable from it after 48. In both studies the history was informative only while the current
state was not yet informative about the target; this reading was formed after the results.
\end{abstract}

\section{Introduction}
\label{sec:intro}

How well a network will learn from further training is not a function of its current loss alone. Networks
lose the ability to fit new targets under continued training \citep{lyle2023understanding,dohare2024loss},
warm-started networks generalise worse than freshly initialised ones \citep{ash2020warm}, and what happens
early in training, such as a sensory deficit or the presence of regularisation, has lasting effects
\citep{achille2019critical,golatkar2019time}. The path to a state matters for what follows.

What these findings leave open is whether the path carries information that the state no longer shows.
The question matters for understanding learning: if the effect of the past is fully stored in the
weights and the optimiser state, then for predicting what comes next the training process needs no
memory beyond its state. It also matters in practice. A monitor that reads a checkpoint can be applied to
any stored model, while a monitor that needs the history has to be carried along the whole run. A
checkpoint can also be probed: a short burst of further training on a chosen task is itself a measurement
of the state, and it may capture what a summary of the history would add.

We test the question on two targets that lie at different distances in the future. The near target is the
response of a network to further training, read by such a probe at a checkpoint. The far target is the
error a run reaches at its end, forecast after a growing part of the run has been observed. For the near
target we fixed a threshold in advance:

\begin{quote}
A recurrent state of at most four dimensions, computed from the recorded training history and selected on
validation data, reduces the held-out error in predicting a network's response to further training by at
least 10\% relative to a baseline that sees only the current checkpoint, with a bootstrap lower bound
above zero, and keeps a mean gain of at least 5\% when a whole history regime is held out.
\end{quote}

A negative answer to such a test means little if the probe cannot tell states apart. The main study
therefore had two stages. The probe first had to register a known manipulation and give repeatable
values; only then was the prediction result read. We report both stages in the order in which they
occurred.

The paper contributes a short response probe checked with a positive control and repeated measurements;
a registered test of whether a compact summary of the history predicts the probe beyond the current
state; and a forecasting screen that varies how much of the run the forecaster has seen. Two screens on
image classification are reported in appendix~\ref{app:image}.

\section{Background}
\label{sec:related}

\paragraph{The path to a state.}
Continued training can reduce a network's ability to learn new targets
\citep{lyle2023understanding,dohare2024loss,abbas2023loss,berariu2021study}, with candidate mechanisms that
include saturation and linearisation of units, growth of parameter norms and large target scales
\citep{lyle2024disentangling}, and changes of curvature \citep{lyle2023understanding}. Resetting or
regularising parts of the network counters it \citep{nikishin2022primacy,kumar2025maintaining}. Studies of
the early phase show that later behaviour depends on the first epochs
\citep{frankle2020early,achille2019critical,golatkar2019time,jastrzebski2020breakeven,fort2020deep}. This
work compares training paths and proposes diagnostics of the state. It does not test whether a summary
of the path predicts future learning better than those diagnostics.

\paragraph{Forecasting from a partial run.}
Extrapolating a learning curve from its first part supports early stopping in hyperparameter optimisation
and architecture search
\citep{swersky2014freezethaw,domhan2015speeding,klein2017learning,baker2017accelerating,chandrashekaran2017previous,wistuba2020rank,ru2021speedy,yan2021nasbenchx11,adriaensen2023lcpfn,rakotoarison2024ifbo,kadra2023dpl};
a survey is given by \citet{mohr2024learning}, and \citet{viering2023shape} review curves over the size of
the training set. Predictors also use weights, gradients and other training telemetry
\citep{yamada2016weight,unterthiner2020predicting,naik2026telemetry}, and simple predictors based on early
values can be competitive with learned ones in some compute regimes
\citep{white2021powerful,egele2024discarding}. These methods use the history; they are rarely compared
with a calibrated predictor of the current state under holdouts of whole task families.

\paragraph{Measuring before interpreting a null.}
Repeatability is quantified with the intraclass correlation \citep{shrout1979intraclass,mcgraw1996forming}.
Positive controls and dose series are standard in measurement validation and rare in studies of training
dynamics. We use both to show that the target responds before a null is read, and we fixed criteria and
analysis before the runs, in the spirit of preregistration \citep{nosek2018preregistration}.

\section{Main study: design}
\label{sec:design}

\begin{figure}[t]
\centering
\includegraphics[width=0.98\linewidth]{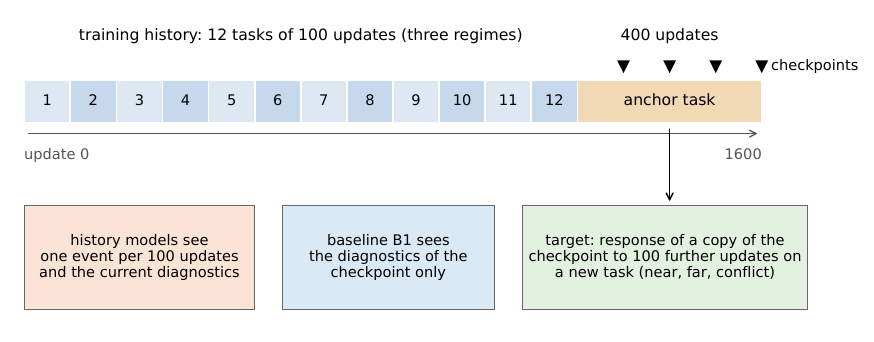}
\caption{Design of the main study. A network is trained on 12 tasks and then on a common anchor task.
At each of four checkpoints, a copy receives 100 updates on a new task and its loss reduction is recorded.
History models see one event per 100 updates of the history together with the diagnostics of the current
checkpoint. The baseline sees the diagnostics only.}
\label{fig:design}
\end{figure}

\Cref{fig:design} shows the design. All constants are taken from the configuration frozen before
execution (appendix~\ref{app:repro}).

\subsection{Networks and histories}

Inputs are $x \sim \mathcal{N}(0, I_{32})$. A fixed orthogonal matrix $Q$ maps them to latent coordinates
$u = Qx$, divided into four subspaces of eight dimensions. A task is a unit vector $v$ in latent space
with target $y = \tanh(v^\top u) + \varepsilon$, where $\varepsilon \sim \mathcal{N}(0, 0.05^2)$ during
training and $\varepsilon = 0$ in every evaluation. The network is a multilayer perceptron with two
hidden layers of 128 rectified linear units, initialised after \citet{he2015delving} and trained on half
the mean squared error with stochastic gradient descent (momentum 0.9, batch size 128). The learning rate
\VLearningRate{} was selected on the anchor task alone, without access to any response.

A history consists of 12 tasks of 100 updates each, followed by 400 updates on a common anchor task;
optimiser state is carried through. Three regimes differ in the 12 tasks: three directions are cycled
(repetitive, HF-REP), every task has a new direction (diverse, HF-DIV), or directions alternate between
cosine $+0.95$ and $-0.95$ with the anchor (conflicting, HF-CON). The task sequence of a regime is drawn
once and shared by its 12 histories, which differ in initialisation and training batches. Six further
histories, the re-initialisation histories, follow the repetitive regime and re-initialise the 25 least
active units of each hidden layer every 400 updates; they serve only as a control of the probe. Histories
are split by seed into 24 for fitting, 6 for model selection and 6 for testing. The unit of all
resampling is the history.

\subsection{The probe}

Checkpoints are taken after 1300, 1400, 1500 and 1600 updates, all within the anchor phase. At a
checkpoint, a copy of the network is trained for $K = 100$ updates on a challenge task, a task that is not
part of the history, with a new optimiser state. The loss $L_k$ on a fixed noise-free batch of 512 inputs
gives the response
\begin{equation}
Y = \frac{1}{K}\sum_{k=1}^{K} \left(L_0 - L_k\right).
\end{equation}
The response measures how the current state reacts to further training, not where the run will end.
Challenges come in three classes defined by their cosine with the anchor direction: near ($0.75$),
conflict ($-0.9$) and far (latent coordinates no history task uses), each with four instances.
Instances 0 to 2 serve fitting and selection, instance 3 is reserved for testing, and the measurement at
instance 0 is repeated with an independent stream of training batches. Responses are standardised per
class on the fitting histories; the three class responses at a checkpoint form a response map.
\VCellsPrelock{} responses were used for fitting and selection, and \VCells{} were reserved and opened
once. The prediction test uses the 72 responses of instance 3 on the six test histories.

\subsection{Predictors}

All predictors share one decoder,
\begin{equation}
\hat{y} = \beta_0 + \beta_x^\top x + \beta_q^\top q + x^\top W_{xq}\, q + \beta_h^\top h + h^\top W_{hq}\, q ,
\end{equation}
where $x$ holds measurements of the current checkpoint, $q$ describes the challenge and $h$ represents the
history. The baseline B1 has no $h$. Its 748 features include training age, current losses including the
loss on the challenge before any update, weight norms, activation statistics, ranks, gradient norms and
cosines, momentum statistics, the loss change of one trial update and the outputs on 64 fixed reference
inputs; it is fit by ridge regression and represents the calibrated current state. The compact state
reads the history as a sequence of events, one per 100 updates, each with the task direction, the loss at
the start and end of the block and phase indicators; a linear recurrence $s_t = A s_{t-1} + P e_t$ with
spectral radius below one gives a state of dimension $d \in \{1, 2, 4\}$, chosen on validation data. A
gated recurrent unit \citep{cho2014learning} with 32 hidden units over the same events is the reference
without the dimension limit. Neither the regime label nor any response is an input. Two further
baselines, a nonlinear model of the current state and a model of the task construction, are described in
appendix~\ref{app:details}.

\subsection{Checking the probe}
\label{sec:control}

For a rectified linear unit, multiplying the incoming weights and bias by $\gamma > 0$ and dividing the
outgoing weights by $\gamma$ leaves the function of the network unchanged and changes the curvature and
the response to gradient steps \citep{neyshabur2015pathsgd,dinh2017sharp}. The positive control applies
this rescaling to a fixed quarter of the units of each hidden layer at the final checkpoint and measures
the response map $R_\gamma$. Its effect is
\begin{equation}
z = \frac{\lVert R_\gamma - R_A \rVert}{\sqrt{3}} - \frac{\lVert R_A - R_B \rVert}{\sqrt{3}},
\end{equation}
where $R_A$ and $R_B$ are maps of the unchanged checkpoint with the same and with an independent stream
of training batches, so that the second term is the noise floor. The design used $\gamma \in \{0.5, 2\}$
on the six test histories. The re-initialisation histories provide a second control, a change of the
state rather than of its parametrisation.

\subsection{Criteria and protocol}

Relative gain is $1 - \mathrm{RMSE}_{\text{model}} / \mathrm{RMSE}_{\text{B1}}$ on the standardised
responses of the test histories. Two validity criteria come first: the intraclass correlation ICC(1,1)
between first and repeated measurement is at least 0.80 in each class, and the positive control reaches a
median effect of at least 0.25, positive in at least five of six histories and in every regime. The
prediction criteria follow: the compact state gains at least 10\% over B1 with a 90\% bootstrap lower bound
above zero, and keeps a mean gain of at least 5\% when one regime is held out. \Cref{tab:gates} lists all
nine criteria in their fixed order; the first criterion not met determines the recorded outcome.
Intervals are percentile intervals from 2{,}000 bootstrap resamples of histories that repeat the model
selection \citep{efron1993introduction}.

The configuration, criteria and analysis were frozen in local version control before the production
run, without an external timestamp; the configuration labels the evidence as exploratory. After the first
result, four further steps were each written down and committed before they were carried out: an
analysis of the first validity results on recorded data; a second stage of the measurement, executed
once; a third repeat of the conflict measurement; and a regularised summary of the full history, fit
without reading test values. The first two changed doses, window and repeats of the measurement and left
the prediction criteria, the fitted models and their predictions untouched. Appendix~\ref{app:protocol} lists the
deviations.

\section{Main study: results}
\label{sec:results}

All results in this section concern the task generator, the network and the \VHistories{} histories
described above.

\subsection{Stage 1: the probe before revision}

The first question was whether the probe could be trusted. The near and far responses were repeatable
(ICC \VIccNear{} and \VIccFar), but the conflict response reached only \VIccConflict{} \VIccConflictInt,
below 0.80. The positive control preserved the function and stayed below its threshold
(\VCtrlOrigHalf{} at $\gamma = 0.5$, \VCtrlOrigTwo{} at $\gamma = 2$). By the fixed order, the hypothesis
was recorded as neither supported nor rejected. The prediction criteria had been computed in the same
pass (\cref{tab:gates}), but a probe that does not register a known manipulation cannot show the absence
of an effect.

\begin{table}[!b]
\centering
\small
\caption{Criteria of the main study in the fixed order. Below the line: the positive control in the
second stage and the declared third repeat. Intervals are 90\% bootstrap intervals over histories.}
\label{tab:gates}
\begin{tabular}{@{}p{0.27\linewidth}p{0.31\linewidth}p{0.27\linewidth}l@{}}
\toprule
Criterion, in the fixed order & Requirement & Result & Outcome\\
\midrule
Known answer & exact within $10^{-12}$ & reproduced & met\\
Repeatability, minimum over classes & ICC(1,1) $\geq 0.80$ & \VIccMin{} \VIccMinInt{} & not met\\
Gain of the compact state over B1 & $\geq 10\%$, lower bound $>0$ & \VGainState\% \VGainStateInt{} & not met\\
Transfer to a held-out regime & mean gain $\geq 5\%$, each regime $>0$ & mean \VLofoMean\% & not met\\
Against the flexible baseline & RMSE difference, lower bound $>0$ & \VFlexDiff{} \VFlexDiffInt{} & not met\\
Dimension of the response map & stable under resampling & three components, stable & met\\
Margin over task geometry & $\geq 2$ points, lower bound $>0$ & \VGainBgamma{} points \VGainBgammaInt{} & not met\\
Compression & state recovers $\geq 70\%$ of the gain of the history GRU & no gain of the GRU; $d^\ast=\VKstar$ & not met\\
Positive control, original doses & median $z \geq 0.25$, $\geq 5$ of 6 positive, each regime positive & \VCtrlOrigHalf{} \VCtrlOrigHalfInt{}, \VCtrlOrigHalfPos{} at $\gamma=0.5$ & not met\\
\midrule
Stage 2: positive control, strongest dose & median $z \geq 0.25$, $\geq 10$ of 12 positive, monotone in dose & \VCtrlEighth{} \VCtrlEighthInt{}, \VCtrlEighthPos{} at $\gamma=0.125$ & met\\
Extension: repeatability, conflict class, third repeat & ICC(1,3) of the mean, lower bound $>0.80$, declared before execution & \VIccThree{} \VIccThreeInt{} & met\\
\bottomrule
\end{tabular}

\end{table}

\subsection{Stage 2: dose response and repeats}
\label{sec:revision}

The recorded data showed why. The conflict response was noisier between repeats than the other classes
while its spread between checkpoints was similar, so averaging repeats addresses it, and the control
effect depended on dose and on the averaging window (\cref{fig:control}a,b; details in appendix~\ref{app:details}).
The second stage was frozen before any new response was generated: doses $\gamma \in \{0.5, 0.25,
0.125\}$, 12 histories, a window ending at step \VPrefixEnd{} read from the recorded curves, and the
unchanged threshold of 0.25. All its criteria were met (\cref{fig:control}c). The median effect rose from
\VCtrlHalf{} to \VCtrlQuarter{} to \VCtrlEighth{} \VCtrlEighthInt, with \VCtrlEighthPos{} histories
positive at the strongest dose. A third repeat of the conflict measurement, declared separately, gave a
reliability of the three-repeat mean of \VIccThree{} \VIccThreeInt. The window and doses differ from those
of the prediction target, the response over 100 updates.

\begin{figure}[t]
\centering
\includegraphics[width=\linewidth]{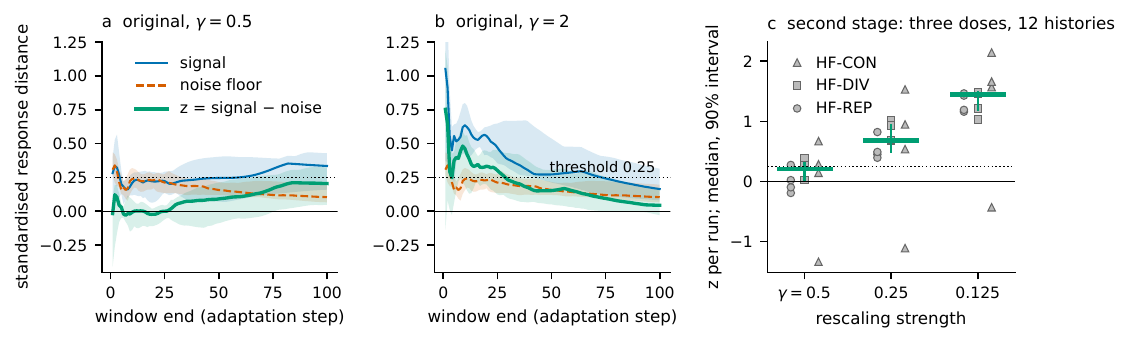}
\caption{Positive control in the main study. a, b: original design on six histories. Lines are medians of
the response distance to the unchanged checkpoint (signal), of the distance between two unchanged
measurements (noise floor) and of their difference $z$, for averaging windows ending at each step; bands
are 90\% bootstrap intervals. c: second stage with stronger doses on 12 histories, window ending at step
\VPrefixEnd. Points are histories, bars are medians with 90\% intervals. The dotted line is the threshold
fixed before the first run.}
\label{fig:control}
\end{figure}

The second control asked whether the probe sees a change of the state itself. Before production, a
frozen technical control compared six pairs of histories that differed only in the re-initialisations,
at the final checkpoint directly after the last one; the probe separated the pairs in every class, by
\VPcrejRatioNear, \VPcrejRatioFar{} and \VPcrejRatioConflict{} times the noise floor between repeats. For the
production histories we registered, before computing it, that the probe would separate the six
re-initialisation histories from the twelve unmanipulated ones at all four checkpoints (effect of the
same form, with the spread between unmanipulated histories as the floor, threshold 0.25). It did at the
final checkpoint (\VPcrejZSixteen, \VPcrejZSixteenInt) and at step 1500 (\VPcrejZFifteen,
\VPcrejZFifteenInt), and not at steps 1300 and 1400 (\VPcrejZThirteen{} and \VPcrejZFourteen), 100 and 200
updates after the previous re-initialisation. The registered prediction was not met; the smaller effect
at step 1500 does not follow the order of distances and is not explained.

\subsection{The history did not improve on the current state}

\begin{figure}[t]
\centering
\includegraphics[width=\linewidth]{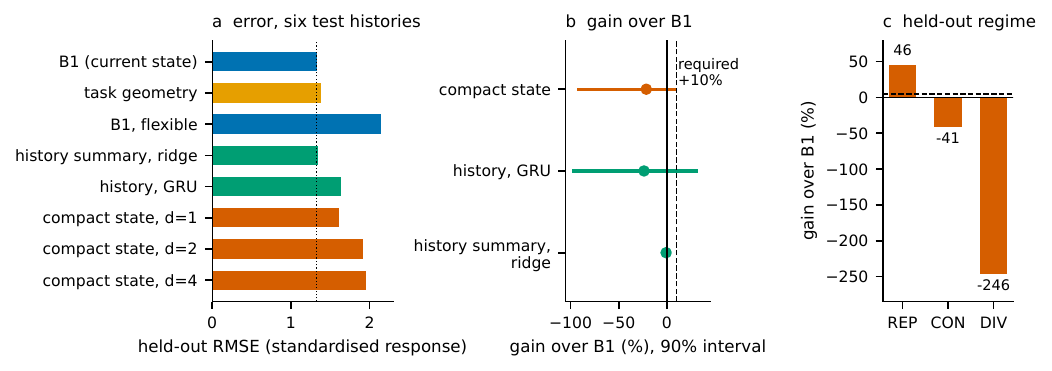}
\caption{Prediction of the response to further training in the main study. a: error on the six test
histories; the dotted line marks the current-state baseline. b: gain over the baseline with 90\% bootstrap
intervals over histories; the dashed line is the gain required in advance. c: gain of the compact state
when a whole regime is held out; the dashed line is the required mean gain.}
\label{fig:prediction}
\end{figure}

With the probe checked, the prediction result can be read. It was computed once in the first analysis
and not changed afterwards (\cref{fig:prediction}). On the six test histories B1 had an error of
\VRmseBone{} (RMSE of the standardised response), the compact state \VRmseState{} and the history GRU
\VRmseHfull. The gain of the compact state was \VGainState\% \VGainStateInt; the upper end of the
interval lies below the required 10\%. The gain of the GRU was \VGainHfull\% \VGainHfullInt. The selected
state dimension was $d^\ast = \VKstar$. With one regime held out, the mean gain was \VLofoMean\%, positive
only for the repetitive regime.

A history model can fail because the history is uninformative or because the model overfits. The GRU
fitted the training histories better than B1 (\VFitTrainHfull{} against \VFitTrainBone) and was worse on
validation (\VFitValHfull{} against \VFitValBone), which points to overfitting with 24 training
histories. The declared extension therefore replaced the recurrent model by a fixed summary of the
history with ridge regression. It matched B1 on validation and stayed slightly worse on test (gain
\VGainShrunk\%, \VGainShrunkInt); because it shares its penalty and checkpoint features with B1, this is a
check that the summary adds nothing, not an independent test. With six test histories the study can
resolve only large gains: detecting a gain of 10\% with 80\% power would need between \VPowerTenE{} and
\VPowerTenT{} test histories (appendix~\ref{app:details}).

\section{Forecasting screen: a target further away}
\label{sec:companions}

The main study asked about the near future. The screen asks the same question about the end of a run, and
lets the amount of observed history grow. It was run in the same project with the same comparison
logic, a model of the history against a model of the current state.

\paragraph{Design.} Thirteen families of scalar functions on $[-1,1]^{32}$ (linear, polynomial, Fourier,
radial basis, piecewise, compositional, and solutions of ordinary and partial differential equations)
were each learned by 120 small networks that vary in architecture, optimiser (AdamW
\citep{loshchilov2019decoupled}, stochastic gradient descent with Nesterov momentum, LARS
\citep{you2017large}), batch size, learning rate and weight decay, giving \MTrajectories{} runs of up to
240 epochs. The target is the logarithm of the normalised root mean squared test error at the end of the
run, on data that never enter the trajectory. After 12, 24 and 48 epochs five forecasters predict it: the
current validation error taken as the forecast, a snapshot model of the latest row of training
telemetry, a compact recurrent state, a GRU and a Transformer \citep{vaswani2017attention} over the
trajectory. Every run of one family is held out in turn and the forecasters are trained on 1{,}000 runs of
the other twelve. The analysis plan, frozen before any forecast existed, set two thresholds at epoch 48:
a mean improvement of at least 10\% of some history model over the snapshot model, and an improvement in
at least 9 of 13 held-out families. The comparison over epochs was planned without a threshold, and the
comparison with the current validation error was planned as context. Intervals are 90\% bootstrap
intervals over the 13 families, without correction for multiple comparisons.

\begin{figure}[t]
\centering
\includegraphics[width=0.97\linewidth]{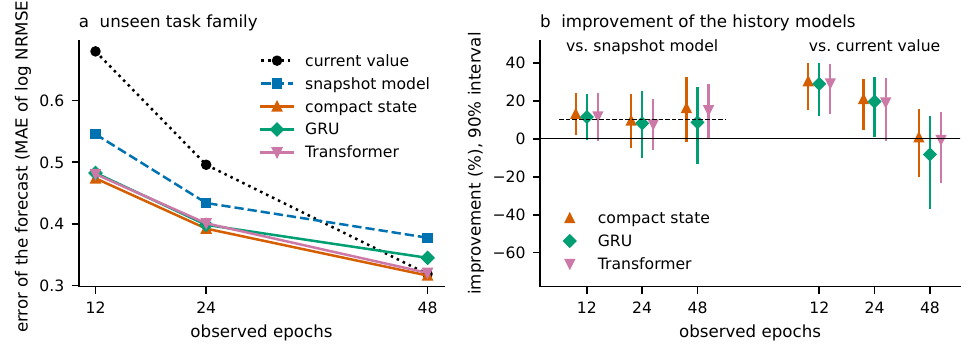}
\caption{Forecasting screen with 1{,}000 runs for training the forecasters. a: forecast error with one task
family held out. b: improvement of the three history models over the snapshot model (left) and over the
current validation error (right), with 90\% bootstrap intervals over families. The dashed line is the
threshold set for epoch 48 against the snapshot model.}
\label{fig:meta13}
\end{figure}

\paragraph{Results.} Early in the run, history helped (\cref{tab:meta13}, \cref{fig:meta13}). After 12
epochs the compact state improved on the snapshot model by \MStateBoneTwelve\% \MStateBoneTwelveInt{}
and on the current validation error by \MStateCurTwelve\% \MStateCurTwelveInt; the GRU and the
Transformer also improved on the current value. After 24 epochs the compact state still improved on the
current value (\MStateCurTwentyFour\%, \MStateCurTwentyFourInt). After 48 epochs the first registered
threshold was met in the mean (Transformer \MFullBoneFortyEight\%, \MFullBoneFortyEightInt, over the
snapshot model) and the second was not (\MFamilyWins{} families). At the same epoch no history model was
distinguishable from the current validation error (compact state \MStateCurFortyEight\%,
\MStateCurFortyEightInt), while the snapshot model was worse than it; the registered gain therefore
reflects in part a weak snapshot model. Results with fewer training runs and without held-out families
are given in appendix~\ref{app:screen}.

\begin{table}[t]
\centering
\small
\caption{Forecasting screen: mean absolute error of the forecast of the final log test error, with one
task family held out (average over 13 families). Bold marks the lowest error in a column.}
\label{tab:meta13}
\begin{tabular}{@{}lrrrrrr@{}}
\toprule
 & \multicolumn{3}{c}{1{,}000 runs for training the forecaster} & \multicolumn{3}{c}{200 runs}\\
\cmidrule(lr){2-4}\cmidrule(l){5-7}
Predictor & epoch 12 & 24 & 48 & 12 & 24 & 48\\
\midrule
Current value, unfitted & 0.680 & 0.496 & 0.319 & 0.680 & 0.496 & \textbf{0.319}\\
Snapshot model & 0.545 & 0.434 & 0.378 & 0.629 & 0.551 & 0.488\\
Compact recurrent state & \textbf{0.474} & \textbf{0.392} & \textbf{0.316} & \textbf{0.483} & \textbf{0.398} & 0.368\\
GRU over history & 0.483 & 0.398 & 0.345 & 0.509 & 0.412 & 0.370\\
Transformer over history & 0.481 & 0.401 & 0.321 & 0.533 & 0.434 & 0.385\\
\bottomrule
\end{tabular}

\end{table}

\section{Discussion}
\label{sec:discussion}

\paragraph{The near target.} We asked whether a compact summary of the history predicts how a network
responds to further training better than its current state. In the main study it did not: the 90\%
interval of the gain lies below the 10\% fixed in advance, the unrestricted GRU overfitted, and a
regularised summary added nothing. The answer can be read because the probe was checked first. It
registered a function-preserving rescaling with a monotone dose response, the mean of three repeats was
reliable, and it saw a re-initialisation of units directly after it. The same check shows its limit: 100
to 200 updates after a re-initialisation, the probe no longer separated those histories.

\paragraph{The far target.} For the end of a run, history helped while little of the run had been seen.
After a fifth of the run the current validation error was as good as any history model. One reading
connects both studies: the history is informative while the current state is not yet informative about
the target, as for the final error after a few epochs, and adds nothing when the state already reflects
it, as for the response to the next 100 updates. This reading was formed after the results. The two
screens on image classification (appendix~\ref{app:image}) found no advantage of history encoders, with weak
evidence, and do not test it.

\paragraph{The path and the state.} In the tested settings the past acted through the state it left
behind. A well-measured state, including a short probe of its response, carried what the history added;
a summary of the path added no predictive value beyond a careful measurement of the present. The value of
the history depended on the target and on timing. It helped early in a run and for a distant target, and
the trace of an intervention faded from the probe within a few hundred updates. In these studies, how far
ahead the target lay and how recent the events were decided whether a record of the past added anything
to a look at the present.

\paragraph{What the paper does not claim.} It does not claim that the current state is sufficient in
general, that the history carries no information, or that a small set of scalar diagnostics would
suffice. It makes no statement about large models or language modelling.

\paragraph{Future work.} The main study shares one task sequence within each regime and probes at least
100 updates after the last history task; task sequences drawn per history and probes directly after a
history task would test the claim more broadly and in a more favourable case. Resolving a gain of 5 to
10\% needs between \VPowerTenE{} and \VPowerFiveT{} test histories. The early window of the forecasting
screen should be tested as a registered hypothesis with correction for multiple comparisons, and the
reading that connects both studies needs a design in which the distance of the target is varied within one
system.

\section{Limitations}
\label{sec:limits}

\begin{enumerate}
\item \emph{Sample size.} The prediction test rests on six test histories and three task sequences; the
forecasting screen on 13 designed families.
\item \emph{Order and registration.} The second stage of the measurement was completed after the first
prediction results were known; its plan was fixed before new data and changed no prediction criterion,
fitted model or prediction. Protocols were frozen in local version control, not registered externally.
The combined evidence is exploratory.
\item \emph{Reliability.} The frozen criterion concerned a single measurement and was not met for one
class; the value above the criterion is for the mean of three repeats.
\item \emph{Scope.} Small networks, synthetic or small image tasks and short horizons; post hoc elements
of the screens are marked in \cref{tab:status}.
\end{enumerate}

\section{Conclusion}

A compact summary of the training history did not predict a network's response to further training
better than a calibrated model of its current state, read through a probe that had first been shown to
register known manipulations. When forecasting the end of a run, the history helped only while little of
the run had been observed. A reading formed after the results connects both: in the tested settings, the
history was informative only while the current state was not yet informative about the target.

\section*{Data and code availability}

The frozen protocols, the analysis code, the result records from which every number and figure of this
paper is generated, and the generating script are available from the authors. Appendix~\ref{app:repro} lists
commits and file hashes.

\section*{Author contributions}

M.H.\ conceived the study, wrote the protocols and the code, ran and analysed the experiments and wrote
the manuscript. P.M.\ supervised the work and revised the manuscript.

\section*{Use of AI-assisted tools}

AI coding and writing assistants (Claude, Anthropic) assisted with research code, run orchestration,
analysis scripts, literature lookup and drafting. The authors are responsible for the design,
verification, interpretation, citations and final text.

\bibliographystyle{plainnat}
\bibliography{references}

\clearpage
\appendix

\section{Protocol history and deviations}
\label{app:protocol}

\paragraph{Main study.} The production data come from protocol version 3, the production instance; no data
of earlier versions are used. The final analysis was computed directly from the completed run records
with the frozen analysis definitions; it used all \VHistories{} histories and all \VCells{} reserved
responses. The steps after the first result were committed before execution and each was executed once.
The second stage of the measurement differs from the original control in dose, number of histories and
averaging window.

\paragraph{CIFAR-100 screen.} The trace includes, for 40 fixed images, the gradient of a logit statistic
with respect to the input. For 20 of the \SRuns{} runs, which had not completed a first update, the
validity check on this gradient was set to record a gradient of exactly zero as zero before they were
executed; each was executed once. Data, architectures, training and analysis are the same for all runs.
The calibrated baselines were added after the first complete analysis.

\paragraph{Forecasting screen.} The reported version uses a final test set that is never observed during
training; an earlier configuration without this separation is not used. Cells in which a secondary
metric overflows are reported as overflow, without clipping or rerun (\MOverflow{} cells).

\section{Further details of the main study}
\label{app:details}

\paragraph{Criteria.} The nine criteria in their fixed order are: known answer (the response computation
reproduces an analytic case), repeatability, gain, transfer to a held-out regime, a nonlinear baseline
(B1 with a hidden layer of 32 units, which the compact state must beat), the stability of the dimension of
the response map, a margin of two percentage points over a task-geometry baseline (B1 with summaries of
the known history tasks relative to the challenge), compression (the compact state recovers at least
70\% of the gain of the GRU) and the positive control. Function preservation of the control is verified
on five input sets to an output tolerance of $10^{-6}$, and a random perturbation of equal norm serves as
a comparison; its norm matched to a relative error of $\VCtrlNormErr$.

\paragraph{Fitting.} Neural predictors are fit with Adam \citep{kingma2015adam} on a fixed grid of two
learning rates and three weight decays, with early stopping on the validation histories, and their
predictions are averaged over five initialisations. The state dimension is the smallest within one
standard error of the best validation error. All choices use fitting and validation histories only, and
no model was refit after the test values were opened.

\paragraph{Stage 1 diagnosis.} In the conflict class the variance between checkpoints was \VConfBetween{}
and between repeats \VConfWithin, against \VNearWithin{} and \VFarWithin{} between repeats for near and
far. At $\gamma = 0.5$ the control effect was positive in \VCtrlOrigHalfPos{} histories, with a largest
value over windows of \VCtrlHalfPeak{} at step \VCtrlHalfPeakStep; at $\gamma = 2$ it was \VCtrlTwoEarly{}
after the first update and \VCtrlTwoLate{} at step 100. The noise floor fell over the window (change
\VNoiseChange, \VNoiseChangeInt). In the second stage, \VCtrlHalfPos, \VCtrlQuarterPos{} and
\VCtrlEighthPos{} histories were positive at the three doses (intervals \VCtrlHalfInt, \VCtrlQuarterInt,
\VCtrlEighthInt), and responses over the shorter window were standardised with the transformation of the
full window.

\begin{figure}[h]
\centering
\includegraphics[width=0.62\linewidth]{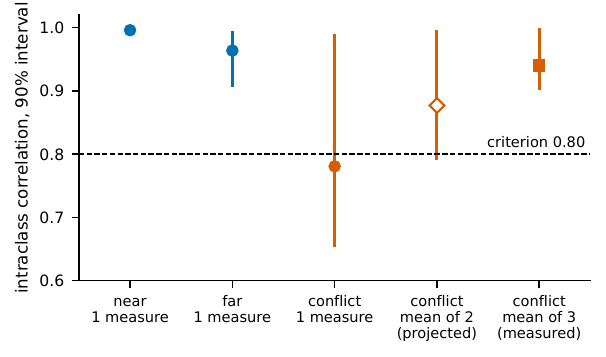}
\caption{Repeatability of the response in the main study, with 90\% bootstrap intervals over histories.
The single conflict measurement missed the criterion. The value for the mean of two repeats is a
projection; the value for three repeats was measured afterwards.}
\label{fig:reliability}
\end{figure}

\paragraph{Reliability.} The mean of two repeats has a projected reliability of \VIccTwoProj{} by the
Spearman--Brown formula, with lower bound \VIccTwoProjLo; this projection is not a pass of the frozen
single-measurement criterion. The third repeat measured \VIccThree{} \VIccThreeInt{} on
\VIccThreeUnits{} checkpoints (\cref{fig:reliability}).

\paragraph{Further baselines.} The nonlinear baseline had an error of \VRmseFlex, worse than B1, so the
criterion comparing it with the compact state is uninformative. The task-geometry baseline had
\VRmseBgamma, between B1 and the history models. States of dimension 2 and 4 had errors of \VRmseStwo{}
and \VRmseSfour. B1 had \VFitTrainBone{} on the fitting and \VFitValBone{} on the validation histories,
and a reduced reference model without checkpoint diagnostics had \VRmseBzero{} on the test histories.
Adding the regime label to B1 left its error unchanged (gain \VFamLabelGain\%). With one regime held out
the gains were \VLofoRep\% (repetitive), \VLofoCon\% (conflicting) and \VLofoDiv\% (diverse). The
regularised summary uses the mean of each event feature, the difference between last and first event and
nine summaries of the diagnostic trajectory; it had \VShrunkVal{} on validation and \VShrunkTest{} on test.

\paragraph{Power.} A calculation on the observed paired differences indicates that detecting a true gain
of 10\% with 80\% power would need between \VPowerTenE{} and \VPowerTenT{} test histories, and a gain of
5\% between \VPowerFiveE{} and \VPowerFiveT, depending on the approximation.

\paragraph{Diagnostic collisions.} A supporting check searched for pairs of checkpoints from different
regimes with nearly equal current-state features (Mahalanobis distance with a shrinkage covariance
\citep{ledoit2004wellconditioned}), nearly equal predicted responses and clearly different measured
responses; at least eight such pairs were required. Of \VCollPairs{} eligible pairs, \VCollBaseline{}
were close in current-state features; all but \VCollPredicted{} of these were excluded because their
predicted responses differed, and the measured responses of the remaining pair were close.

\paragraph{Status of the findings.} \Cref{tab:status} labels each finding by how it was obtained.

\begin{table}[h]
\centering
\small
\caption{Status of the findings.}
\label{tab:status}
\begin{tabular}{@{}p{0.5\linewidth}p{0.22\linewidth}p{0.2\linewidth}@{}}
\toprule
Finding & Obtained & Units\\
\midrule
Main study: no gain of the compact state or the history GRU & fixed in advance & 6 test histories\\
Main study: no transfer to a held-out regime & fixed in advance & 3 regimes\\
Main study, stage 1: conflict response below the reliability criterion & fixed in advance & 24 checkpoints\\
Main study, stage 2: dose response of the positive control & fixed before new data, after the first
result & 12 histories\\
Main study: reliability of the three-repeat mean & extension declared before execution & 24 checkpoints\\
Main study: re-initialisation control, paired & technical control frozen before production & 6 pairs\\
Main study: re-initialisation control in production & registered after the first result on existing
data; prediction not met & 18 histories\\
Main study: no gain of the regularised history summary & extension after the first test result, fit
without reading test values & 6 test histories\\
Main study: collision coverage, power & post hoc on recorded data & \\
Forecasting: thresholds at epoch 48 & fixed in advance & 13 families\\
Forecasting: advantage over the snapshot model at epoch 12 & planned profile, no threshold & 13 families\\
Forecasting: comparison with the current value & reported as context, no decision rule & 13 families\\
CIFAR-100: reservoir against calibrated current accuracy & design fixed; baseline and reported arm post
hoc & 3 blocks\\
Class-incremental: forecasters compared & fixed in advance & 4 class orders\\
\bottomrule
\end{tabular}
\end{table}

\section{Further results of the forecasting screen}
\label{app:screen}

With 200 training runs the history models keep their advantage at epochs 12 and 24 and are worse than the
current value at epoch 48. With 100 training runs the snapshot and Transformer models produced extreme
forecasts in two held-out families (mean error at epoch 48 of \MHundredBone{} for the snapshot model
against \MHundredState{} for the compact state). When runs of all families are available for training,
the compact state is no better than the snapshot model (\MInterpStateTwelve{} against
\MInterpBoneTwelve{} at epoch 12, \MInterpStateFortyEight{} against \MInterpBoneFortyEight{} at epoch 48).
The families are a designed set, and the intervals describe stability across this set. \MAtCap{} runs
reached the epoch limit, so the end point is a fixed stopping rule and not convergence. The forecasters
are not matched in capacity.

\section{Reproducibility}
\label{app:repro}

Result numbers in the text are macros written by \texttt{scripts/render\_assets.py} from the result
records in \texttt{sources/}; the tables of results and all figures are produced by the same script. The
file \texttt{generated/provenance.json} lists the SHA-256 of each source file and the source of each
macro. Design constants and the count of 20 runs in appendix~\ref{app:protocol} are typed. The script
\texttt{scripts/audit\_paper.py} checks that the hashes match, that every macro used is defined, that
every citation has an entry, that no placeholder is left, and that decimal numbers typed in the text
belong to a list of design constants.

Main study: main study protocol v3.0, global seed 20260827; SHA-256 of the frozen configuration (first 16
digits): \texttt{\VConfigHash}. Second stage of the measurement at commit \texttt{\VCommitRevision};
extensions at commit \texttt{9e91fc51145d}. SHA-256 of the file of analysis values (first 16 digits):
\texttt{\VValuesHash}. Forecasting screen: commit \texttt{ccb5e357e126}; bootstrap seed 13032026 with
20{,}000 draws. CIFAR-100 screen: SHA-256 of the analysis record
\breakhash{abb6e617612e51f99f4d6992fc67ad119f4277f27b7dd63b0f956638cf89a795}. Class-incremental screen:
training commit \texttt{cfe34b811363}, analysis commit \texttt{7d2f13fda442}.

\section{Two screens on CIFAR-100}
\label{app:image}

\subsection{CIFAR-100 subsets with two architectures}
\label{sec:cifar}

\paragraph{Design.} Twelve classification tasks of 20 classes each were built from CIFAR-100
\citep{krizhevsky2009cifar} in three blocks of four. A ResNet-18 \citep{he2016deep} and a VGG-11
\citep{simonyan2015very} were trained on each task with three seeds for 60 epochs, \SRuns{} runs. The
target is the test accuracy after epoch 60, forecast after 6, 12 and 24 epochs from a trace of weight
spectra, training telemetry and a representation sketch. History encoders are fixed reservoirs of 64
units (an echo state network and four spiking variants;
\citealp{jaeger2001echo,maass2002realtime,lukosevicius2009reservoir}) with a linear or kernel readout, ten arms in
total. In the main evaluation a block of tasks is held out. The design was fixed before the runs and set
no numerical thresholds. With three blocks there is no estimate of uncertainty; all values are point
estimates. Baselines that calibrate the current validation accuracy on the training folds were added
after the first analysis, when a check showed that the unfitted current accuracy was too weak a
comparison. The reported reservoir is the best of the ten arms on the test folds. Both were added after
the first analysis and are marked as such in \cref{tab:status}.

\begin{figure}[h]
\centering
\includegraphics[width=0.47\linewidth]{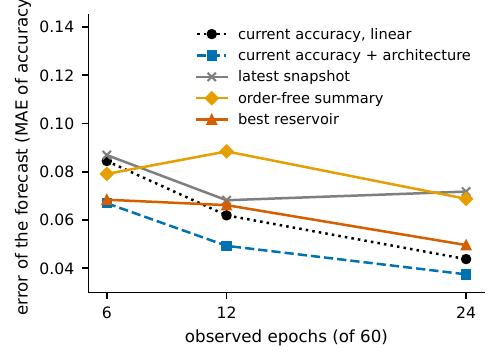}\hfill
\includegraphics[width=0.47\linewidth]{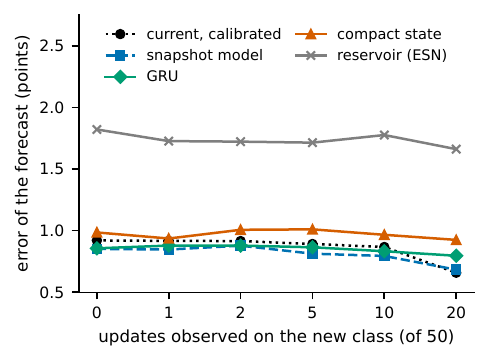}
\caption{Left: CIFAR-100 screen, forecast error with a block of tasks held out; the unfitted current
accuracy (error \SCurSix{} to \SCurTwentyFour) is off the scale. Right: class-incremental screen, mean
forecast error over five targets in percentage points with a class order held out.}
\label{fig:cifarinc}
\end{figure}

\paragraph{Results.} The best reservoir had an error of \SBestSix, \SBestTwelve{} and \SBestTwentyFour{}
after 6, 12 and 24 epochs (\cref{tab:cifar} and \cref{fig:cifarinc}, left). A quadratic function of current accuracy with
architecture terms had \SQuadSix, \SQuadTwelve{} and \SQuadTwentyFour. The reservoir was thus no better
than this predictor at any horizon, by a small margin at epoch 6 and a larger one later. It was better
than the linear calibration of current accuracy at epoch 6 and worse at epochs 12 and 24. It
was better than an order-free summary of the same trace in all six comparisons; order controls did not
isolate an effect of order.

\paragraph{Limits.} Three held-out blocks, tasks from one image corpus, reservoir settings fixed and not
tuned, and a post hoc baseline.

\begin{table}[h]
\centering
\small
\caption{CIFAR-100 screen: mean absolute error of the forecast of final test accuracy (as a fraction)
with a block of tasks held out. $\dagger$: selected or added after the first analysis. For the snapshot,
summary and reservoir rows, the better of the available readouts on the test folds is shown for each
column.}
\label{tab:cifar}
\begin{tabular}{@{}lrrr@{}}
\toprule
Predictor & epoch 6 & epoch 12 & epoch 24\\
\midrule
Current accuracy, unfitted & 0.2865 & 0.2201 & 0.1363\\
Current accuracy, linear calibration & 0.0844 & 0.0619 & 0.0437\\
Current accuracy and architecture, quadratic$^\dagger$ & 0.0669 & 0.0492 & 0.0374\\
Latest multichannel snapshot & 0.0869 & 0.0681 & 0.0717\\
Order-free summary of the prefix & 0.0791 & 0.0884 & 0.0687\\
Reservoir, best of ten arms$^\dagger$ & 0.0684 & 0.0661 & 0.0496\\
\bottomrule
\end{tabular}

\end{table}

\subsection{Class-incremental CIFAR-100}
\label{sec:incremental}

\paragraph{Design.} A ResNet-18 learned the 100 classes one at a time with 50 updates per class, under
seven regimes: plain fine-tuning, a lower learning rate, replay with 20 or 100 stored examples per class
\citep{rebuffi2017icarl}, distillation from the previous model \citep{li2018learning}, online elastic
weight consolidation \citep{kirkpatrick2017overcoming,schwarz2018progress}, and replay with distillation.
With four class orders and two seeds this gives \IRuns{} runs; the setting is class-incremental in the
sense of \citet{vandeven2022three}. For each new class, five quantities at the end of its 50 updates are
forecast (accuracy on the new class, on old classes and on all seen classes, forgetting, and an
area measure of how quickly the class is learned) after 0, 1, 2, 5, 10 and 20 of the updates. The design,
including the correction for multiple tests \citep{holm1979simple}, was fixed before the runs. A complete
class order is held out, so there are four units for inference.

\paragraph{Results.} We report the input set that contains the current errors and the training
telemetry, one of four input sets of the design (\cref{tab:incremental} and \cref{fig:cifarinc},
right). In this input set the snapshot model, which sees only the current state, had the lowest mean
error up to 10 updates and the calibrated current values at 20. The GRU was better than the calibrated
current values up to 10 updates, by at most \IGruMaxGain{} points, and worse at 20; it was not better
than the snapshot model at any horizon. The compact state was worse than the calibrated current values at
every horizon, and the echo state network was worse than them in \IEsnWorse{} combinations of input set,
horizon and target. No difference survived the correction for multiple tests.

\paragraph{Limits.} The outcomes are nearly constant. Over \IObservations{} class stages, most new-class
accuracies are 100\% and most old-class accuracies are zero, because the network predicts the newest class
for almost every input; replay delays this collapse and does not prevent it. A low forecast error
therefore reflects a predictable pattern, and the screen has little power to separate forecasters. The
unfitted current values have errors of \ICurrentFirst{} to \ICurrentLast{} points, so most of the
accuracy of all fitted forecasters comes from calibration.

\begin{table}[h]
\centering
\small
\caption{Class-incremental screen: mean absolute forecast error in percentage points, averaged over five
targets, after $u$ updates on the new class, for the input set with current errors and training
telemetry. Bold marks the lowest error in a column.}
\label{tab:incremental}
\begin{tabular}{@{}lrrrrrr@{}}
\toprule
Predictor & $u=0$ & 1 & 2 & 5 & 10 & 20\\
\midrule
Current values, calibrated & 0.920 & 0.916 & 0.915 & 0.891 & 0.867 & \textbf{0.658}\\
Snapshot model & \textbf{0.851} & \textbf{0.846} & \textbf{0.878} & \textbf{0.813} & \textbf{0.794} & 0.683\\
Order-free summary of the prefix & 2.372 & 1.868 & 1.796 & 1.249 & 1.846 & 1.308\\
Echo state network & 1.821 & 1.727 & 1.722 & 1.714 & 1.776 & 1.661\\
Compact recurrent state & 0.985 & 0.936 & 1.006 & 1.010 & 0.966 & 0.925\\
GRU over history & 0.856 & 0.878 & 0.878 & 0.865 & 0.832 & 0.796\\
\bottomrule
\end{tabular}

\end{table}

\end{document}